\documentclass[runningheads]{llncs}

\usepackage{graphicx}
\usepackage{float}
\begin{document}

\raggedbottom

\mainmatter  

\title{Outlining the design space of eXplainable swarm (xSwarm): experts’ perspective}

\titlerunning{Explainable Swarm}

%
%
\author{Mohammad Naiseh\inst{1} \and
    Mohammad D. Soorati\inst{2} \and
    Sarvapali Ramchurn\inst{2}}
\authorrunning{Naiseh et al.}
%

\institute{Department of Computing and Informatics, Bournemouth University, United Kingdom \and Department of Electronics and Computer Science, University of Southampton, United Kingdom\newline
\email{mnaiseh1@bournemouth.ac.uk}\\
\email{\{m.soorati, sdr1\}@soton.ac.uk}
}
\maketitle

\begin{abstract}
In swarm robotics, agents interact through local roles to solve complex tasks beyond an individual’s ability. Even though swarms are capable of carrying out some operations without the need for human intervention, many safety-critical applications still call for human operators to control and monitor the swarm. There are novel challenges to effective Human-Swarm Interaction (HSI) that are only beginning to be addressed. Explainability is one factor that can facilitate effective and trustworthy HSI and improve the overall performance of Human-Swarm team. Explainability was studied across various Human-AI domains, such as Human-Robot Interaction and Human-Centered ML. However, it is still ambiguous whether explanations studied in Human-AI literature would be beneficial in Human-Swarm research and development. Furthermore, the literature lacks foundational research on the prerequisites for explainability requirements in swarm robotics, i.e., what kind of questions an explainable swarm is expected to answer, and what types of explanations a swarm is expected to generate. By surveying 26 swarm experts, we seek to answer these questions and identify challenges experts faced to generate explanations in Human-Swarm environments. Our work contributes insights into defining a new area of research of eXplainable Swarm (xSwarm) which looks at how explainability can be implemented and developed in swarm systems. This paper opens the discussion on xSwarm and paves the way for more research in the field.

\end{abstract}

\section{Introduction}
Swarm robotics consists of multiple robots that interact with each other to accomplish a collective task that is beyond individual ability, or can be performed more effectively by a group of robots ~\cite{divband2021designing}. Swarm robotics promise various implementations in many scenarios such as pollution control, surveillance, package delivery and firefighting. Although swarms can perform the task autonomously, human intervention is still crucial for safe and effective deployment in real-world scenarios ~\cite{agrawal2021explaining}. For example, swarm operators may need to manage swarm tasks that might be beyond swarm capability or intervene in cases of failures and errors. This intervention could be more complex when it requires multiple humans to work together in a team to operate remote swarms ~\cite{clark2022industry}. For instance, in a search and rescue scenario where the swarm is responsible for searching and identifying casualties, the system may consist of two operators: (a) an operator is tasked with scheduling clusters of robots to search areas and (b) an analyst who is tasked with analysing casualties images. 

Human-Swarm Interaction (HSI) is an emerging field of research that studies human factors and swarm robotics to better utilise both humans and swarm capabilities in a Human-Swarm environment. An essential part of HSI is communicating, to humans, the reasons behind swarms’ decisions and behaviour ~\cite{roundtree2019transparency}. As swarms become more advanced, explaining their behaviour (individual and collective) is buried in increasingly complex communication between swarm agents and underlying black-box machine learning models. 

Explainable Artificial Intelligence (XAI) is a response to this direction of re-search. XAI is a domain of research that is interested in making the behaviour or the output of intelligent systems more intelligible to humans by providing explanations~\cite{miller2019explanation}. While the concept of XAI is increasingly gaining attention in several research domains, such as Agent Planning ~\cite{anjomshoae2019explainable}, Machine Learning ~\cite{naiseh2021explainable} and Motion Planning ~\cite{brandao2021experts}. There is a clear gap in the literature on integrating swarm robotics into the research and development of XAI. Similar to XAI fields of research, we argue that eXplainable Swarm (xSwarm) should address that an explanation is an answer to an unexpected event or a failure of the system~\cite{miller2019explanation}. However, compared to different XAI domains, xSwarm can be significantly different since the decision made by a swarm significantly depends on the communication and collective contribution of its agents ~\cite{roundtree2019transparency}. Further, the explainability of swarms would require explaining the emerging collective behaviour that is not necessary for the summation of the individual robot behaviour ~\cite{brambilla2013swarm}. 

Many studies discussed the need for XAI methods and models for Human-Swarm environments to improve trust and maintain situational awareness, e.g., ~\cite{mualla2022quest,roundtree2019transparency,agrawal2021explaining}. However, at the current stage of research, it is still vague what kind of explanations a swarm system should be expected to generate and what questions an explainable swarm is expected to answer. This paper seeks to conceptualise the concept of xSwarm and introduce a new taxonomy of explainability requirements for HSI. Our results can guide the requirements elicitation for explanation methods and models for Human-Swarm teaming. These insights contribute to operationalising the concept of xSwarm in reality and can serve as a resource for researchers and developers interested in designing explainable Human-Swarm interfaces. 

In this paper, we limit our research to swarm robotics, i.e., this paper does not address multi-robot systems that have explicit goals and their agents execute both individual and group plans. We follow a similar approach proposed in recent publications to outline explainability requirements in a given problem~\cite{liao2020questioning,brandao2021experts}. Our method uses open-text and scenario-based questionnaire with domain experts, to elicit explainability considerations for swarm systems. We foresee a lack of established understanding of the concept or common technical knowledge given the early stage of xSwarm in industrial practices. Therefore, we used a novel use-case developed in our recent work ~\cite{clark2022industry} and supported with failure scenarios and explanations to ground our investigation. Our data set was collected from on an online questionnaire with 26 swarm experts. In this paper, we make the following contributions: 
\begin{itemize}
  \item We present xSwarm taxonomy that summarises the explainability space for swarm robotics from swarm experts' perspective.
  \item We summarize current challenges faced by researchers and industry practitioners to create xSwarm.
\end{itemize}

\section{Background and related work}

So far, the process of generating explanations for AI systems users has received a lot of attention from multiple domains, including psychology ~\cite{mueller2019explanation}, artificial intelligence~\cite{lundberg2020local}, social sciences~\cite{miller2019explanation} and law~\cite{atkinson2020explanation}. Furthermore, regulatory requirements and increasing customer expectations for responsible artificial intelligence  raise new research questions on how to best design meaningful and understandable explanations for a wide range of AI users~\cite{naiseh2020personalising}. To respond to this trend, many explainable AI (XAI) models and algorithms have been proposed to explain black-box ML models ~\cite{naiseh2020personalising}, agent-path planning~\cite{anjomshoae2019explainable}, multi-agent path planning~\cite{almagor2020explainable} and explainable scheduling~\cite{kraus2020ai}. These explanations can range from local explanations that provide a rationale behind a single decision to global explanations that provide explanations for the overall AI logic~\cite{lundberg2020local}. 

Despite a growing interest in explainable AI, there is a limited number of studies on how to incorporate explainability in swarm robotics. Only a few attempts in the literature have discussed explainability in swarm robotics. For instance, a scalable human-swarm interaction has been proposed to allow a single human operator to monitor the state of the swarm and create tasks for the swarm~\cite{divband2021designing}. Their study with 100 participants showed that users have different visualisation preferences for explanations to observe the state of the swarm depending on the type of the given tasks. The explanation in this study was limited to a visual representation of the swarm coverage using a heatmap or individual points space that does not capture the wider space of xSwarm. Another example is the work proposed in ~\cite{mualla2022quest}. They proposed Human-Agent EXplainability Architecture (HAEXA) to filter explanations generated by swarm agents with the aim to reduce human cognitive load. Finally, Roundtree et al.~\cite{roundtree2019transparency} discussed design challenges and recommendations for explainable Human-Swarm interfaces, e.g., they described the delivery of explanations to human operators in swarm systems as the main challenge due to human processing limitations caused by a large number of individual robots in the swarm. Although these studies provide steps toward the right direction for xSwarm, it is still unclear whether explanations developed in other XAI domains would be useful for Human-Swarm environments. Also, it is still undefined what kind of questions a swarm system is expected to answer or what explanations from swarm robotics would look like.

Meanwhile, recent papers were published in multiple domains to guide the exploration of explainability requirements in a certain domain. Liao et al.~\cite{liao2020questioning} created a list of prototypical user questions to guide the design of human-centred explainable black-box ML. Similarly, Brandao et al.~\cite{brandao2021experts} proposed a taxonomy of explanation types and objects for motion planning based on interviewing motion planning experts and understanding their current work and practice. Our work is driven by a similar approach for outlining xSwarm's design space. More particularly, in light of the growing interest in XAI, we are interested in understanding what is required to create xSwarm and what are the major practical challenges to developing xSwarm.

\section{Background and related work}

Ethical approval was acquired via the Ethics Committee of the School of Electronics and Engineering from the University of Southampton [Reference No. 67669]. 

\subsection{Participants}
Thirty-two participants voluntarily accessed our online questionnaire. Participants were approached through two main mailing lists: multi-agent\footnote[1]{Multi-agent systems | The Alan Turing Institute\newline  https://www.turing.ac.uk/research/interest-groups/multi-agent-systems} and TAS-Hub\footnote[2]{UKRI Trustworthy Autonomous Systems Hub (TAS-Hub): https://www.tas.ac.uk/} mailing lists. Four participants were excluded because they identified themselves to have less than one year of working in swarm robotics. Two more participants were also excluded as they chose not to complete the questionnaire, leaving us with twenty-six experts. 46.2\% of our participants were researchers, 7.7\% were lecturers, 23.1\% were PhD students, and 22.9\% with industrial roles. All experts had at least more than one year of working in swarm robotics. 61.5\% of participants had 1-5 years of experience, 23.1\% had 5-10 years, 7.7\% with 10-15 years and 7.7\% exceeded 15 years of experience in the field. 

\subsection{Study procedure and design}

Due to the limited research in xSwarm, we chose to conduct an online questionnaire as it allows us to scope an early investigation into outlining a preliminary xSwarm taxonomy for future in-depth investigations. It also helps us to gather a large number of experts’ feedback in a brief time compared with other data collection methods such as interviews~\cite{mcguirk2016using}. We designed our questionnaire to include several elicitation probes to obtain representative results and minimise cognitive heuristics and bias~\cite{morgan2014use,klein1989critical} (Appendix A - Table A.1). First, a draft questionnaire was developed by the first author. Then, the draft was pilot tested with experts (researchers from the inside and outside the research group) to assure that question formulation can achieve the research goal and be understandable to study participants. The questionnaire starts by asking participants to read the participant information sheet. The information sheet outlines the motivation of the study, compiles background material about the research goal and problem as well as describes the questionnaire structure. Participants’ information sheet also contains background material about explainable AI to minimise any availability bias~\cite{morgan2014use}. Then participants were directed toward the main questionnaire. The questionnaire has four main sections. Table 1 provides a brief description of the questionnaire structure and probe types used in each section.  

\begin{table}[H]
\caption{Questionnaire structure and elicitation probes}
\label{tab2}
\scalebox{1}{
\begin{tabular}{|p{3.5cm}| p{2.5cm}| p{6cm}|}
\hline
\textbf{Questionnaire section} &\textbf{Probe type}&\textbf{Description}\\
\hline
Swarm experience&NA&Demographic questions\\
\hline
Swarm failures&Knowledge&Participants’ knowledge about swarm failures, e.g., What are the main reasons for swarm failures? And What behaviour sparks your concern?\\
\hline
xSwarm scenario&Situation assessment&Participants’ assessment of a swarm failure or unexpected behaviour, e.g., What questions would you ask to debug this failure? Why do you think the swarm failed in this case?\\
\hline
xSwarm challenges&Experience&Participants’ previous experience in generating explanations from a swarm, e.g., What kind of challenges did you face when generating an explanation from a swarm?\\
\cline{2-3}
{}&Hypothetical&Participant needs to develop eXplainable swarms, e.g., How would swarm developers be supported in integrating explainability??\\
\hline
\end{tabular}
}
\end{table}

\textbf{Swarm experience.} In this section, we asked participants to provide four demographic factors related to their previous swarm experience - \emph{participants’ area of research or experience, current role, years of experience in swarm robotics, and level of experience in swarm robotics}. We excluded participants who had less than a year of experience or had identified themselves with a low level of experience in the field. 

\textbf{xSwarm scenario.} This section used an industrial-based use case for HSI, which represents a real-world use case for HSI where explainability is a crucial requirement~\cite{clark2022industry}. The use case presented the following scenario: “\emph{A hurricane has affected a settlement. Many were evacuated, but some did not manage to escape and have been caught in debris and fallen structures. A UAV (Unmanned Aerial Vehicle) swarm is being deployed to identify locations of survivors, to evaluate their condition, and to advise on where unmanned ground res-cue vehicles should be deployed to evacuate casualties.}” A diagram for the use case can be found in Appendix A – Figure A.1. First, we asked participants to select when and what the swarm should explain its behaviour to human operators during the task. Then, we provided two cases of swarm failures or unexpected behaviour and asked participants to explain these cases. This is a situation assessment probe to determine the bases for explanations during failures or unexpected behaviour, rather than waiting for these incidents to happen~\cite{klein1989critical}. We also asked participants to provide what questions will they ask to debug failures. Fig. 1.A and Figure 1.B show two scenarios of either a failure on an expected behaviour during two main stages of the swarm task: (a) Planning: a swarm has suddenly stopped while it is trying to reach its destination, and (b) Searching: the swarm has failed to identify all the casualties in the search area. 

\begin{figure}
    \centering
    \includegraphics[width=1.0 \textwidth]{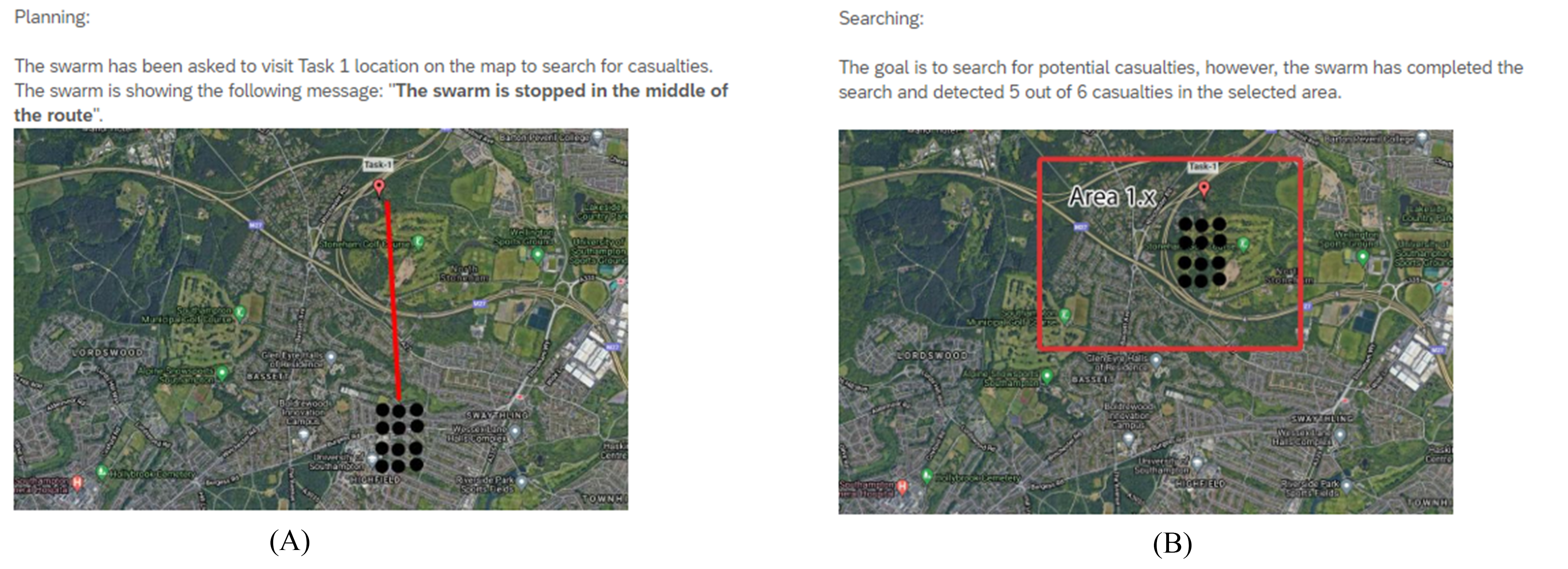}
    \caption{Scenarios of swarm failures or unexpected behaviour presented to participants (A) swarm has suddenly stopped. (B) Swarm has failed to recognise all possible casualties in the search area. }
    \label{fig:pcm}
\end{figure}

 \textbf{xSwarm challenges.} The questions in this section aim to understand the challenges faced by experts to generate explanations from swarm systems. First, this section included open-end questions to reflect on participants’ experience in generating or obtaining explanations from swarm systems. We asked questions such as \emph{“What were the typical difficulties you encountered while trying to extract explanations from swarm systems?”}. The section also contained hypothetical questions to identify the key tools and techniques that swarm developers should have to build xSwarm. We asked questions such as \emph{“How would swarm developers be better supported in integrating explainability?”} and \emph{“What kind of issues and considerations should developers have in mind when developing xSwarms?”}.
 
\subsection{Analysis}

We followed the content analysis method~\cite{elo2008qualitative}. Our data set included participants’ answers to open-ended questions with themes extracted across perspectives. The analysis was conducted by the first author. To increase the trustworthiness of our qualitative analysis, we followed an iterative process across the research team. This iterative procedure was implemented to check and confirm that the taxonomy is reflected in the responses of the participants. The authors met multiple times during each stage of the research to ensure the accurate interpretation of each category and the supporting evidence. These discussions and meetings resulted in the dividing, altering, removing, or adding of categories to ensure that all responses and their settings were adequately represented in the data.

\section{Results}

Our results are divided into two parts. We start by discussing the general explanation categories emergent in participants’ answers, which outline the design space of xSwarm. We then discuss the main challenges faced by our experts to integrate explainability in swarm systems.  Table 2. shows a sample of participants’ answers. More examples can be found in Appendix A - Table A.2.  

\begin{table}[]
    \caption{Explanation categories for swarm robotics. \emph{Explain} column refers to participants' explanations to swarm failures or unexpected behaviour- \emph{Debug} column presents participants' questions to debug swarm failures or unexpected behaviour - \emph{Freq.} refers to the number of times this category appeared in the data set.}
    \centering
    \resizebox{\textwidth}{!}{\begin{tabular}{|p{3cm}|p{7cm}| p{7cm}|}
    \hline
         \textbf{Category}  & \textbf{Explain}  & \textbf{Debug} \\  \hline
        \textbf{Consensus}  & Detecting the missing casualty caused a conflict between the agents. \newline Agents were not able to agree on some features of the casualty.
        \newline \textbf{Freq.= 24}
     &
    How much of the swarm is consolidating the same view?  \newline
    What is the agreement percentage between agents?\newline\textbf{Freq.= 18}\\  \hline
        \textbf{Path Planning}  & Because the swarm initiative plan is to visit Location L which is a charging location. \newline
Because swarm is trying to avoid obstacles\newline\textbf{Freq.= 22}
 & Is it a collision-free route? \newline
Where is the swarm right now?  [the probability distribution of possible robots locations]\newline\textbf{Freq.= 14}\\ \hline
        \textbf{Communication}  & Because the communication is limited between agents. \newline
Environmental condition is limiting the connection between agents
  \newline\textbf{Freq.= 19}& What is the accuracy of information exchange between swarm agents? \newline
What is the synchronisation status between swarm agents?\newline\textbf{Freq.= 6}  \\ \hline
        \textbf{Scheduling}  & Because this casualty should be detected by another cluster. \newline
Because there are not enough agents to detect all casualties \newline\textbf{Freq.= 14}& Why there is x number of agents to search this area? \newline
Why a cluster c1 is assigned to task t1?\newline\textbf{Freq.= 8} \\ \hline
        \textbf{Hardware}  & Because the casualty is out of sensors coverage distance.  \newline
Because of low-quality input data from swarm sensors\newline\textbf{Freq.= 16} & What is the communication bandwidth between swarm robots. \newline
What is the limitation of the swarm sensors?\newline\textbf{Freq.= 5} \\ \hline
        \textbf{Architecture and design}  & Because robots have limited communication links per robot. \newline
Because the swarm has not got enough training in such an environment.\newline\textbf{Freq.= 5}
 & 
What is the response threshold function? \newline
How is the swarm size selected?\newline\textbf{Freq.= 6} \\ \hline
    \end{tabular}}
    \label{tab:Explanation_categories}
\end{table}

\subsection{Explanation categories}

\subsubsection{Consensus.} A key distinction between swarms and multi-agent systems is that the swarm can be seen as a single entity rather than multiple individual robots with multiple goals and tasks. Swarm robotics is usually equipped with methods to reach a consensus on the swarm decision, based on the accumulation and sharing of information about features of the decision encountered by individual robots. The most frequently asked questions by our participants were not only regarding why the swarm has made a single decision but at a high level, acquiring who and how many robots contributed to that decision. Our participants frequently asked questions to check either the percentage of individual robots who communicated the preferred decision, e.g., P18\footnote[3]{For anonymity, individual participants are referred to as Pn (e.g. P18) in this chapter.} mentioned: \emph{“How many robots detected the casualties?”}, and P26 described: \emph{“Which features of the casualty did the robots agree on?”}. Furthermore, when participants were asked to provide reasons for swarm failures, they frequently answered that such a failure could be related to a conflict between swarm agents. Participants enquired information to further understand the opinion dynamics of the swarm. Opinion dynamics is a branch of statistical physics that studies the process of agreement in large populations of interacting agents~\cite{poli2007particle}. Because of these dynamics, a population of agents might progressively transition from a situation in which they may hold varying opinions to one in which they all hold the same opinion. Participants were interested in how the swarm opinion dynamics might change and evolve. For instance, P20 reflected on the swarm searching for casualties task: \emph{“What we are also interested in is that What are the emerging decisions (fragmentation), Is there a dominant decision, how dominant is this decision, and how will this opinion evolve”.} 

\subsubsection{Path Planning.} Participants frequently asked questions related to the path that the swarm is going to follow and its related events. For instance, P2 mentioned that remote human operators would frequently check what is the next state of the swarm. Similarly, P20 added, \emph{“automatic planning has not been fully adopted especially in high-risk domains such this scenario, what the system usually does to suggest the plan to the human operator to check that path”}. Further, participants were not only interested to validate the swarm plan but also asked questions related to the path planning algorithm to debug swarm failures, e.g., P18 and P7 asked to debug a failure in Fig. 1.A: \emph{“What metrics does the swarm use to follow this route?”} and \emph{“Why did the swarm follow this route, not another one?”}. A similar pattern appeared in participants’ answers when they were asked to explain the unexpected behaviour in Fig. 1.A Participants explained the unexpected behaviour based on path planning events, P9 stated: \emph{“Because the swarm agents are charging their batteries according to the initial plan”}, similarly P18 commented: \emph{“Probably the swarm has not stopped, it is just avoiding obstacles right now”}. These results support the growing interest in Explainable Planning (XAIP), as shown by many planning contributions such as Human-Aware Planning and Model Reconciliation~\cite{chakraborti2017plan} and multi-agent path planning~\cite{almagor2020explainable}. Such explanations aim to answer human operators’ questions regarding the path for several agents or clusters of agents to reach their targets~\cite{almagor2020explainable}, such that paths can be taken simultaneously without the agents colliding. However, the research on explainable swarm path-planning is still limited and requires further attention.  

\subsubsection{Communication.} The collective decision-making process is the outcome of an emergent phenomenon that follows localised interactions among agents yielding a global information flow. As a consequence, to investigate the dynamics of collective decision-making, participants went beyond debugging consensus between agents and asked questions related to the communication between swarm agents, e.g., P7 pointed out the following question: \emph{“What is the synchronisation status between swarm agents?”.} Participants also brought up that swarm failure in our examples could be critically related to a failure in ensuring the appropriate flow of behavioural information among agents, both in terms of information quantity and accuracy, e.g., P21 commented on the failure in Fig. 1.B: \emph{“because the environmental condition is limiting the connection between agents”} and P18 added: \emph{“Perhaps one of the casualties moved on / no longer is, therefore state was not updated”}.  

\subsubsection{Scheduling-based.} Task scheduling in swarm robotics is a collective behaviour in which robots distribute themselves over different tasks. The goal is to maximize the performance of the swarm by letting the robots dynamically choose which task to perform.  Yet, in many applications, human operators are still required to intervene in scheduling plan and adapt accordingly. In our questionnaire, participants’ understanding of particular swarm failure was sometimes associated with an initial task scheduling. Questions and explanations related to task scheduling were mentioned in both scenarios. For instance, P2 and P21 explained casualty detection failure (Fig. 1.B) with the following explanations, \emph{“Because this casualty should be detected by another cluster”}, and \emph{“Because the initial scheduling plan did not allocate enough robots”}.   Interestingly, regardless the failure scenarios, our participants repeatedly suggested questions with a pattern proposed by Miller~\cite{miller2018explanation} which have a contrastive nature of explanation that are often implying why not another scheduling plan is feasible. For instance, P19 asked the following question: \emph{“Why is a cluster c1 better than cluster c2 in performing task t1?”}. Our data also showed a set of questions that require interactive scheduling-based explanations, where the human operator can understand the consequences of the scheduling plan. This pattern is pointed out by Wang et al.~\cite{wang2014also} as what-if questions, P12 mentioned: \emph{“What if cluster c1 were to do task t1 rather than task t2?”}.  

\subsubsection{Hardware.} Even though the quality and robustness of the hardware are increasing, hardware failure is still quite common. Participants explained the failure of the swarm with a link to a hardware problem. For instance, when the swarm was not able to detect all the casualties in the area, P17 commented: \emph{“because of faulty in the sensors”}. Our participants also discussed that explanations of hardware are necessary to give human operators actionable explanations to either better utilise the swarm or to improve the swarm performance, e.g., send more swarm agents to the location. There was also a typical pattern among participants’ feedback to explain swarm behaviour or failure based on its hardware boundaries and limitation as well as environmental and training data limitations. For instance, P11 explained unexpected behaviour in Figure 4.1: \emph{“Because some agents have limited speed capabilities, so they are waiting for each other”}, and similarly, P14 explained Figure 4.2 failure, \emph{“the swarm sensors have low image input”}.  

\subsubsection{Architecture and design.} Participants also found many reasons for swarm failures based on the swarm architecture or design decisions of the system. This category did not appear frequently in the data, and participants mentioned five explanations and six questions. Participants recognised that potential failure or unexpected behaviour can be related to initial design decisions made by the systems engineers e.g., P12 answered: \emph{“Because the swarm has not got enough training in such an environment”} and similarly, P22 commented: \emph{“Because the transition probability between states is fixed”}. Participants also took a broader view of swarm explainability when they discussed questions that can be answered through descriptive information about the swarm system design. For instance, P1 and P4 asked: \emph{“How is the swarm size selected?”}, \emph{“What is the design method? is it behaviour-based or Automatic design?.”} Currently, these kinds of explanations are discussed in the XAI literature as an on-boarding activity where human operators are introduced to the system design architecture to understand its capability and limitation~\cite{cai2019hello}. These global properties of the system could help inform the interpretation of swarm behaviour and predictions during critical decision-making scenarios. 

\subsection{Challenges faced by swarm experts }
In this section, we discuss three main challenges participants faced when designing or developing explanations for swarm robotics.  

\subsubsection{A trade-off between explainability and Human-Swarm performance.} Participants discussed that developing explainable swarms comes with a critical analysis of performance-explainability trade-offs that swarm developers have to face. Participants frequently pointed out that generating explanations does not come without cost; they mentioned that this could be a computational or performance cost, e.g., interruption to the human-swarm task. P18 mentioned: \emph{“asking the operator to read an explanation at an emergent situation can be vital and the explanation might delay the reaction time and distract the operator”}. These observations are consistent with earlier research~\cite{naiseh2021explainable} that showed explanations can be ignored by human operators when they are perceived as a task goal impediment. It has been shown that humans in Human-AI systems focus to complete the task rather than trying to engage cognitively with AI systems to understand their behaviour~\cite{buccinca2021trust}. Such effect can be even amplified in human-swarm environments where swarm operators might face time constraints and multitasking as well as large numbers of agents to monitor.  In summary, participants argued that integrating explainability in the Human-Swarm task is difficult – numerous design decisions require trade-offs involving explainability, workload and usability.  

\subsubsection{Increasing the number of agents would increase the complexity of explanation generation}

The xSwarm challenge is significantly more difficult than other XAI areas since a swarm's decision is dependent on numerous parameters linked to numerous agents~\cite{karga2019using}. Participants pointed out that there is a scalability problem in swarm explainability, i.e., increasing the number of agents in a swarm will exponentially increase the complexity of the explanation generation. P12 mentioned: \emph{ “explainable swarm can have an extreme cost when the swarm size is 100”}. Similarly, P2 added: \emph{“some of the technical reasons that led to the decision are relevant to a given user; as the number of agents increases, the non-relevant information increases exponentially”}.  For these reasons, simplicity as a usability feature for explanation can be essential in swarm environments. Simplicity states that explanations generated by AI systems should be selective and succinct enough to avoid overwhelming the explainee with unnecessary information~\cite{sokol2020explainability}.  In response to this challenge, participants suggested that future swarm systems shall include methods and tools to help summarise explanations generated by swarm agents, e.g., P20 suggested: \emph{“Human operators cannot deal with loads of explanations, there should be some tools to provide a summary of the swarm decision”}. In one relevant research, Mualla et al.~\cite{mualla2022quest} addressed this issue and presented a novel explainable swarm architecture with a filtering feature to filter explanations generated from swarm agents. They conduct a user evaluation with three experimental groups: “No explanation: the swarm does not present an explanation”, “Detailed explanation: the swarm presents explanations generated by each agent” and “Filtered explanation, the swarm presented a filtered explanation of each agent”. They showed that filtering explanation was the most preferable option for participants and lead to more trust and understanding of the swarm behaviour.   

\subsubsection{The soundness of swarm explanations is difficult to achieve}

The soundness of an explanation measures how truthful an explanation is with respect to the underlying system state~\cite{sokol2020explainability}. Participants stated that they faced difficulties to generate explanations that are sound and complete because, in some contexts, information might be unavailable or outdated, P7 mentioned: \emph{“A variety of technical factors like bandwidth limitations and swarm size, an environmental condition factors associated with swarms diminish the likelihood of having perfect communication, which negatively impacts generating explanations that are complete and sound”}. To address this challenge, participants suggested explanation generation tools in swarm robotics shall take into consideration the validity of the available information to generate an explanation for a swarm action. P16 mentioned: \emph{“swarm explanation algorithms should have metrics to measure the soundness of their explanations before presenting it to operators”.} Participants also suggested that outdated information about the swarm may result in an outdated explanation which may degrade Human-Swarm performance and trust, P4 raised this issue: \emph{“Sometimes the swarm explanation might be not synchronised with the current state of the swarm, in these cases, it might be better to avoid showing this information to human operators”}. Previous empirical studies discussed this issue and showed that erroneous actions taken by the human operator to progress toward the overall goal will be amass if actions are based on outdated swarm explanations or information~\cite{walker2012neglect}. Future xSwarm research may need to consider discovering mitigation strategies to break the error propagation cycle that may emerge from outdated explanations.   

\section{Conclusion}
Although the research in eXplainable AI is gaining a lot of interest from industry and academia, the research on explainability for swarm robotics is still limited. In this paper, we presented results from an online questionnaire with 26 swarm experts to outline the design space of eXplainable Swarm (xSwarm). In particular, we presented a list of explanations generated by swarm experts to explain swarm behaviours or failures and what kind of questions do they ask to debug the swarm. Our work provides concrete taxonomy of what xSwarm would look like from swarm experts' perspective, serving as an initial artefact to outline the design space of xSwarm. Our work suggests opportunities for the HCI and swarms robotics communities, as well as industry practitioners and academics, to work together to advance the field of xSwarm through translational work and a shared knowledge repository that represent an xSwarm design space. Finally, we cannot claim this is a complete analysis of xSwarm design space. The results only reflect swarm experts’ views using an online questionnaire. Future work could use our taxonomy as a study probe for in-depth investigation with multiple stakeholders in swarm systems to further enhance our taxonomy.    

\section*{Acknowledgments}
This work was supported by the Engineering and Physical Sciences Research Council (EP/V00784X/1).

\bibliographystyle{splncs04}
\bibliography{samplepaper.bib}
\end{document}